\documentclass[A4, 12pt]{article}
\usepackage[top=1in,bottom=1in,left=1in,right=1in]{geometry}

\usepackage[compact]{titlesec}
\usepackage{cite, hyperref}
\usepackage{amsmath,amssymb,amsfonts,enumitem}
\usepackage{algorithm, algcompatible}
\usepackage{graphicx, setspace}
\usepackage{textcomp, comment}
\usepackage{xcolor}
\usepackage{subfigure} 
\newcommand{\X}{\mathbf{X}}
\def\BibTeX{{\rm B\kern-.05em{\sc i\kern-.025em b}\kern-.08em
    T\kern-.1667em\lower.7ex\hbox{E}\kern-.125emX}}
   
\titlespacing*{\section}{0pt}{3pt}{3pt}
\titlespacing*{\subsection}{0pt}{3pt}{3pt}

\begin{document}
\setstretch{1}
\title{Selective Cascade of Residual ExtraTrees }
\author{\small{Qimin Liu}\\
\small{\textit{Department of Psychology and Human Development}} \\
\small{\textit{Vanderbilt University}}\\
\small{Nashville, TN}\\
\small{qimin.liu@vanderbilt.edu}\\
\and
\small{Fang Liu, Ph.D.}\\
\textit{\small{Department of Applied and Computational Mathematics \& Statistics}} \\
\small{\textit{University of Notre Dame}}\\
\small{Notre Dame, IN}\\
\small{fliu2@nd.edu}}
\date{}
\maketitle

\noindent \textbf{Conflicts of Interest}: The authors declare that they have no conflicts of interest.

\begin{abstract}

We propose a novel tree-based ensemble  method named Selective Cascade of Residual ExtraTrees (SCORE). SCORE draws inspiration from representation learning, incorporates regularized regression with variable selection features, and utilizes boosting to improve prediction and reduce generalization errors. We also develop a variable importance measure to increase the explainability of SCORE.   Our computer experiments  show that SCORE provides comparable or superior performance in prediction against ExtraTrees, random forest,  gradient boosting machine, and neural networks; and the proposed variable importance measure for SCORE is comparable to studied benchmark methods. Finally, the predictive performance of SCORE remains stable across hyper-parameter values, suggesting potential robustness to hyper-parameter specification.
\end{abstract}

\noindent \textbf{keywords}: Extremely randomized trees, boosting, ensemble learning, regularized regression, variable importance measure, explainability

\setstretch{1.05}
\section{Introduction}\label{sec:intro}

Ensemble learning methods combine multiple learning algorithms to  achieve better learning performance than that from individual learning algorithms. The success of ensemble methods, in part, can be attributed to the diversity among the  constituent members, which helps mitigate over-fitting and reduce the generalization  error\cite{Dietterich2000AnRandomization}. 

We focus on tree-based ensemble methods for regression. Tree-based ensemble methods construct more than one decision tree and, by appreciating the diversity among the trees, increase the generalizability of the ensemble. Such diversity often comes from perturbation in the optimization process of individual trees. For example, tree bagging generates bootstrap samples, from which decision trees are obtained and averaged \cite{Breiman1996BaggingPredictors}. Random subspace selects a pseudo-random subset of features when building trees \cite{Ho1998TheForests}. Random  forest (RF) selects a random subset of the features for each candidate split, and trains the trees with bootstrap samples \cite{Ho1995RandomForest}. 

The computational costs increase drastically as the number of trees increase in tree-based ensemble approaches given both the optimization and the perturbation processes. One way to reduce the computational burden is to replace optimization with randomization processes. C4.5, for example, randomly chooses from the best 20 splits in constructing individual trees \cite{Dietterich2000AnRandomization}; extremely randomized trees (ExtraTrees) randomly selects cut-point for each candidate split variable \cite{Geurts2006ExtremelyTrees}; and \cite{Fan2003IsEfficiency} proposes  ``randomly'' choosing a non-tested feature at each level of the tree  without using any training data.  

However, the extreme randomness and lack of optimization can lead to prediction bias for  individual tree. \cite{Liu2008SpectrumTrees} suggests that a continuum of tree randomization exists between two extremes: deterministic algorithms with injected randomization and the complete randomization paradigm; and proposes an algorithm to generate a range of models between the two extremes.  Another approach to reduce bias is through gradient boosting machine (GBM) \cite{Friedman2001GreedyMachine,Mason2000BoostingDescent}.  Rather than combining trees independently in a parallel fashion, GBM builds one tree at a time in sequential orders and each new tree aims to predict the residuals from the previous tree (i.e., functional gradient descent). GBM performs better than RF if  parameters are tuned carefully. However,  disadvantages exist for GBM: First, hyperparameters in GBM is arguably harder to tune than RF; Second,  GBM can be prone to overfitting; Third, the  model training can be computationally burdensome given the sequential order in tree construction.

Recent works have emerged in embedding decision trees in complex structures, such as neural networks (NNs) \cite{hinton2006fast}, to further reduce bias. \cite{Kong} stacks the RF onto deep NNs to improve the prediction accuracy in classification for the  $n<p$ case. \cite{Zhou2017DeepNetworks} embeds RF in a NN architecture, referred to as the ``deep forest'' approach. \cite{Feng2018Multi-layeredTrees} adopts a multi-layer architecture where gradient boosting decision trees serve as base learners. With output from prior layers functioning as input for next layers, these methods optimize base learners in each layer by improving the output of the prior layer so that the quality of input increases for the subsequent layers. The effectiveness of these approaches highlights that optimized base learners associate with predictive accuracy for the current layer and serve as useful inputs in the multi-layer context.  

As one of multi-layer learning methods, NNs have gained vast popularity in both machine learning and applied sciences. However, for multi-layer methods with deep structures, difficulty arises in tracing the contributions of the input features. That is, the deep structure can generate black-box predictions that lack interpretability or explainability. The deep structure in NNs also requires a relatively large amount of training data and careful tuning of hyper-parameters. NN models often 1) involve a large of amount of parameters, 2) can be subject to overfitting were inappropriate or insufficient regularizations  employed, 3) rely on high-performance computational resources to yield fast training and learning. For these reasons, NNs do not necessarily offer the best performance for all types of learning tasks on all types of data. Regardless, representation learning, as one of the key properties of NNs, has received wide attention for its role in enhancing predictive performance \cite{bengio2013representation}. We propose to regard decision trees as representation features. Specifically, we develop an ensemble learning approach that incorporates representation learning, sparsity regularization, and boosting, to achieve accurate predictions. 

We term our proposed method Selective Cascade of Residual ExtraTrees  (SCORE).  Our proposed tree-based ensemble method can offer  high prediction accuracy, explainability regardless of deep structures, and low computational costs.  
Methodologically, SCORE incorporates several effective  machine learning strategies, including extremely randomly trees \cite{Geurts2006ExtremelyTrees}, boosting \cite{Freund1997ABoosting, Boosting}, sparsity regularization \cite{Li2009SelectiveFramework}, and representation learning \cite{Bengio2013RepresentationPerspectives}. The reasons for choosing each of the specific learning techniques to build SCORE are provided below.  
\vspace{-6pt}
\begin{itemize}[leftmargin=12pt]\setlength\itemsep{-1pt} 
\item  To maximize the diversity of the base learners and reduce computational costs, SCORE employs the extremely randomized trees (ExtraTrees) as the base learners. In our experiments, the computational cost of SCORE is much lower than the benchmarks methods.
\item To reduce prediction bias in each individual tree in the ensemble and set up a framework to measure ``importance'' of the input features for predictions, SCORE filters out irrelevant trees for prediction via sparsity-regularized regression, referred to as the TreeSelection step. The selected trees  are regarded as representation features. The TreeSelection is a critical step for SCORE and differentiates it  from other tree ensemble approaches. Existing tree ensemble approaches often include irrelevant features in building  trees or constructing layers, which can associate with low  prediction accuracy.  In our experiments, the predictions from SCORE are significantly better in general than its competitors without a tree selection step. 
\item  To further enhance prediction accuracy, we employ the boosting technique by forming multiple layers of ExtraTrees + TreeSelection, where the outcome in each layer is the residuals given the predictions up to that layer. Boosting itself is an ensemble method for improving prediction, by training weak learners sequentially and combining them to form a strong learner. In the context of SCORE, each layer of ExtraTrees + TreeSelection can be regarded as a weak learner.  
\item To improve the explainability of the predictions by SCORE, we develop a conceptually and computationally simple yet effective variable importance measure (VIM). The VIM correctly identifies the top predictors in our experiments.
\end{itemize}
In summary, SCORE leverages the advantages of the existing tree ensemble approaches but improves on computational cost by using a random set of base learners, on outcome prediction via boosting and regularization,  and on explainability with a  VIM. The ensemble is  reflected not only  horizontally through building random tress but also vertically via boosting. In addition, our experiments suggests that SCORE does not require precise tuning on the hyperparameters compared to other state-of-the-art ensemble methods to yield comparable or superior results.

\section{Selective  Cascade of Residual ExtraTrees (SCORE)} \label{s2} 
Let $y$ represent the observed outcome and $\bar{f}$ be the prediction of $y$ from an ensemble method. \cite{Brown2005ManagingEnsembles} decomposes the generalization error $E(\bar{f}-y)^2$ of the ensemble method as follows 
\begin{equation}\label{eqn:ge}
\begin{split}
\textstyle M^{-2}\left[(\sum_i(E(f_i)-y))^2+\sum_i E(f_i-E(f_i))^2+\sum_i \sum_{j\neq i} E\left((f_i-E(f_i))(f_j-E(f_j))\right)\right]
\end{split}
\end{equation}
where $M$ is the number of individual base learners in the ensemble and $f_i$ refers to prediction from the $i$-th learner in the ensemble for $i=1,\ldots, M$. The  terms in Eqn (\ref{eqn:ge}) are respectively the squared average bias, average variance, and average covariance across the ensemble members. The covariance serves as an indicator of the ensemble diversity. The smaller the covariance, the greater diversity there is among the ensemble members. 

\begin{figure}[!htb]
\centerline{\includegraphics[width=1\textwidth]{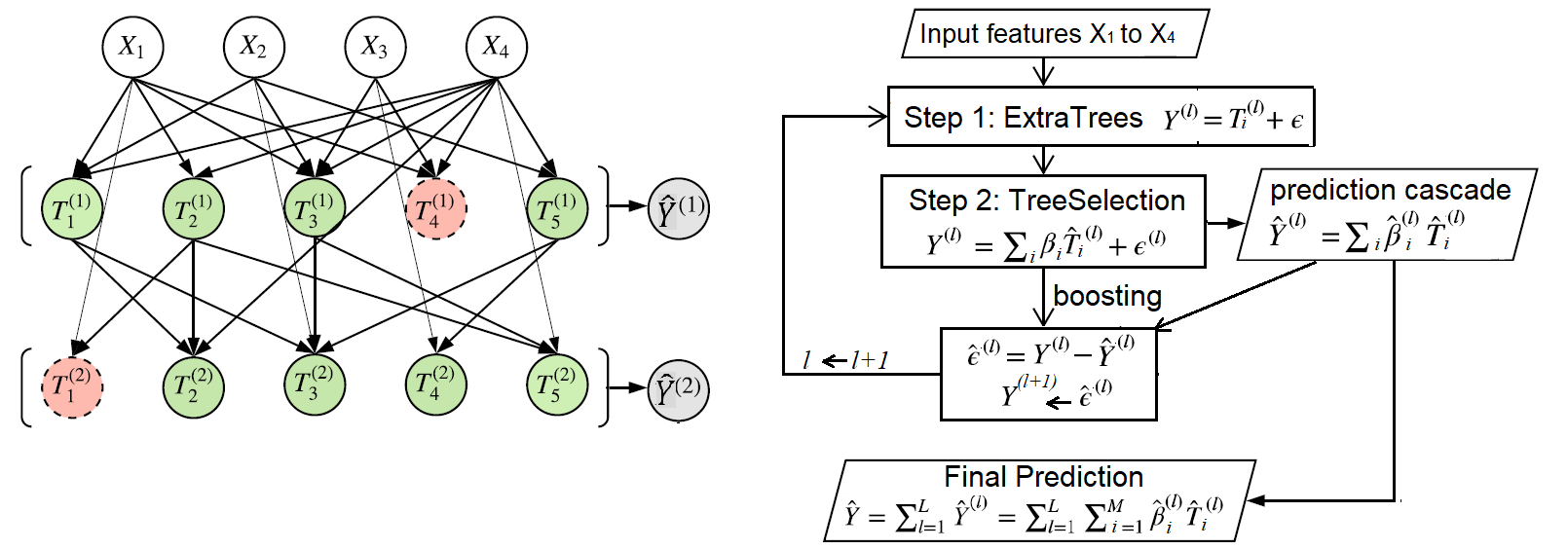}} \vspace{-6pt}
\caption{An example ensemble structure in SCORE and algorithm flowchart (green circles with  solid edges and red circles with dashed edges represent selected and non-selected trees via TreeSelection, respectively.)}
\label{fig1}
\vspace{-6pt}
\end{figure}

Our proposed SCORE procedure aims at 1) maximizing the diversity and thus reducing the covariance term among the ensemble trees  in Eqn (\ref{eqn:ge}) by employing the extremely randomized tree procedure; 2) reducing the mean squared error term (the sum of the squared bias and the variance terms in Eqn (\ref{eqn:ge}) by implementing a selection step via sparsity-promoting regularized regression on individual trees as the representation features; 3) further reducing  the bias and the variance of the ensemble by using weighted averages of predictions from the multi-layered boosting.  Algorithm \ref{alg:SCORE} lists the steps for the SCORE procedure. Figure \ref{fig1} depicts an example of a 2-layer SCORE with 4 original attributes ($X_1$ to $X_4$), along with the algorithmic flowchart. $T_i^{(l)}$ refers to the $i$-th individual randomized tree  constructed in layer $l$ for $l=1,2$. $Y^{(l)}$ is the outcome in layer $l$, which is the residual from the regression in layer $l-1$ for $l>1$ (that is, $Y^{(l)}=Y^{(l-1)}-\hat{Y}^{(l-1)}$), and is the original $Y$ when $l=1$. 
\begin{algorithm}[!htb]
\caption{The SCORE Algorithm}\label{alg:SCORE}
\begin{algorithmic}[1]
\small
\State \textbf{input}: training set $\mathcal{D}_t$; validation set $\mathcal{D}_v$; stopping threshold $\tau_0$
\State \textbf{output}: trained SCORE ensemble leaner and VIM.
\State Employ ExtraTrees to construct an initial set of randomized trees (base learners) on $\mathcal{D}_t$. Calculate the mean squared prediction error MSE$^{(1)}$ on $\mathcal{D}_v$.
\State Let $\Delta^{(1)}$= MSE$^{(1)}$; $l\leftarrow 2$
\WHILE{ $\Delta^{(l-1)}>\tau_0$}
\State \parbox[t]{6in}{TreeSelection: Apply sparsity-regularized regression to outcome $Y^{(l-1)}$ to select relevant representation features  and  calculate residuals $Y^{(l)}$. \strut}
\State \parbox[t]{6in}{ExtraTrees: Employ ExtraTrees to predict $Y^{(l)}$ with either the global-$X$ or local-$X$ input feature option. \strut}
\State\parbox[t]{6in}{Calculate  MSE$^{(l)}$ on $\mathcal{D}_v$, and set $\Delta^{(l)}\!=\!$ MSE$^{(l-1)}\!-$MSE$^{(l)}$.\strut}
\State  $l\leftarrow 1+1$
\ENDWHILE
\State Calculate VIM.
\end{algorithmic}
\end{algorithm}

In what follows, we describe the ExtraTrees and TreeSelection steps of the SCORE procedure in more details, and propose an approach to measure the importance of the original input feature which aids the interpretability of the SCORE procedure and its output.  We use $\X^{(l)}$ to denote the predictor attributes in layer $l$, and provide two options. The first option, referred to as the \emph{local $X$ option}, limits $X^{(l)}$ only to the representation attributes learned from layer $l-1$ and selected by the TreeSelection step (see Sec \ref{sec:treeselection}). The second option, referred to as the \emph{global $X$ option},  allows  $\X^{(l)}$ to include not only the selected representation attributes from layer $l-1$, but also the original input features as well as the learned representation features from all previous layers.

\subsection{Construction of Extremely Randomized Trees}\label{sec:ExtraTrees}
The first step in each layer is the construction of extremely randomized trees as base learners. We employ the ExtraTrees procedure to achieve that goal.  ExtraTrees offers prediction accuracy comparable to RF and higher than tree bagging due to the ``extreme'' randomness. The randomization process in ExtraTrees drastically increases diversity among trees at a lower computational cost compared to RF. The reduced computational costs come from decreased optimization in ExtraTrees as optimization process can be a limiting factor for computation speed. 

ExtraTrees starts by randomly selecting $K$ attributes out of all attributes. Then, ExtraTrees chooses a random cut-point from the observed range of each of the $K$ attributes. After that, ExtraTrees selects the attribute resulting in the smallest MSE as the final split variable. A pre-set minimal node size can be used as a stopping rule for the splitting process.  Denote the number of randomized trees in layer $l$ by $M$ (for notation simplicity, we set $M$ fixed across all layers in SCORE, but $M$ can vary by layers), and let $T_i^{(l)}$ be the $i$-th randomized tree for $i=1,\ldots, M$.  We express the tree-based regression for each tree $T_i^{(l)}$  in layer $l$ as  
\begin{equation}
Y^{(l)}=T_i^{(l)}(\X^{(l)})+\epsilon_i^{(l)},
\end{equation}
where $\epsilon_i^{(l)}$ is the error term, the distribution of which does not need to be specified.

\subsection{Tree Selection via Sparsity Regularization}\label{sec:treeselection}
The predicted outcome $\hat{T}_i^{(l)}$ from each randomized tree can be regarded as a representation feature. Not every representation feature is a good predictor of $Y^{(l)}$, especially given the way how trees are built in ExtraTrees. Including irrelevant features in a regression model can lead to increased bias in prediction and generalization errors. To mitigate this problem, SCORE applies a sparsity-promoting regularization to select more relevant representation features from the $M$ randomized trees. The regularized loss function in layer $l$ is 
\begin{equation}\label{eqn:SCORE}
L\left(Y^{(l)}, \boldsymbol{\beta}^{(l)}\mathbf{\hat{T}}^{(l)}\right)+ \lambda^{(l)} R\left(\boldsymbol{\beta}^{(l)}\right), 
\end{equation}
where $\boldsymbol{\beta}^{(l)}$ quantifies the ``weights'' for the trees in $\mathbf{\hat{T}}^{(l)}=(T_1^{(l)},\ldots,T_M^{(l)})$, and $R(\boldsymbol{\beta}^{(l)})$ is the regularizer on $\boldsymbol{\beta}^{(l)}$ with  tuning parameter $\lambda^{(l)}$. Different types of loss function can be used on $L$, such as $l_1, l_2$ or negative log-likelihood. If the least-squared regression  ($l_2$ loss function) with the $l_1$ regularizer \cite{Tibshirani1991RegressionLasso} is used, then Eqn (\ref{eqn:SCORE}) becomes 
\begin{equation}
\big|\big|Y^{(l)}- \boldsymbol{\beta}^{(l)}\mathbf{\hat{T}}^{(l)}\big|\big|_2^2+\lambda^{(l)}||\boldsymbol{\beta}^{(l)}||_1.\label{eqn:SCOREl}
\end{equation}
Minimizing Eqn (\ref{eqn:SCOREl}) leads to estimate $\hat{\boldsymbol{\beta}}^{(l)}=(\hat{\beta}_1^{(l)},\ldots,\hat{\beta}_1^{(M)})$, some of which are exactly 0 with the $l_1$ regularization. The trees associated with non-zero $\hat{\beta}$ make the set (local $X$ option) or a subset (global $X$ option) of the input features for the next layer. The residual $Y^{(l)}-\hat{Y}^{(l)}= Y^{(l)}-\hat{\boldsymbol{\beta}}^{(l)}\mathbf{\hat{T}}^{(l)}$ serves as the outcome $Y^{(l+1)}$ for the next layer.

\subsection{Cascade Layers and Final Outcome Prediction}
Multiple layers with the dual ExtraTrees+TreeSelection step are built, each with a newly calculated outcome and a new set of input features (newly identified representation tree features from the last layer, plus or not the input features from all previous layers). The layers are stacked on until there is no meaningful improvement on the prediction MSE for the validation set. Suppose there are totally $L$ layers after running the SCORE algorithm, the overall prediction model can be written as
\begin{align}\label{eqn:model}
Y= \textstyle\sum^L_{l=1}\left(\beta_0^{(l)}+\boldsymbol{\beta}^{(l)}\mathbf{T}^{(l)}(\mathbf{X}^{(l)})\right)+\varepsilon^{(L)}.
\end{align} 
Denote the final output prediction by $\hat{Y}$. If each layer is weighted equally, then 
\begin{align}
\hat{Y}=\textstyle\sum^L_{l=1}\left(\hat{\beta}_0^{(l)}+\hat{\boldsymbol{\beta}}^{(l)}\hat{\mathbf{T}}^{(l)}(\mathbf{X}^{(l)})\right).\label{eqn:yhat}
\end{align} 
A generalized version of the SCORE prediction is to apply different weights to different layers, say by incorporating a learning rate $\gamma\in(0,1]$ in a similar manner as in GBM, 
\begin{align}
\hat{Y}=\textstyle\sum^L_{l=1}\gamma^{l}\left(\hat{\beta}_0^{(l)}+\hat{\boldsymbol{\beta}}^{(l)}\mathbf{T}^{(l)}(\mathbf{X}^{(l)})\right).\label{eqn:gamma}
\end{align} 
When $\gamma=1$, it reduces to  equation (\ref{eqn:yhat}). For $\gamma<1$, the deeper a layer is, the less contribution it will have toward the final prediction of $Y$. The weighting scheme can help to mitigate overfitting with an oversized cascade structure.

\subsection{Variable Importance Measure (VIM)}\label{sec:VIM}
The VIM we propose for SCORE is based on the weighted frequency of each variable being selected for node splitting in all the randomized trees, where the weights are the estimated  coefficients $\hat{\boldsymbol{\beta}}$ from the the regularized regression in Eqn (\ref{eqn:SCORE}). If a regression coefficient $\beta^{(l)}$ is associated with the original input features ($l=1$ with local X option; and $l\ge1$ with global X option), then it can be used directly to quantify the contributions of that feature in the outcome prediction. For $\beta^{(l)}$ associated with a representation feature, its value will be traced back, layer by layer, to the original input features used for building these trees. To achieve this, we define two frequency vectors $\mathbf{c}_{X,i}^{(l)}$ and $\mathbf{c}_{T,i}^{(k,l)}$. 

Denote the dimension of the original $\mathbf{X}$ by $p$. Let $\mathbf{c}_{X,i}^{(l)}$ be a column vector of length $p$ for the $i$-th randomized tree in the $l$-th layer. $\mathbf{c}_{X,i}^{(l)}$ contains the frequency on each input attribute used as a split variable in that tree. For example, suppose $\mathbf{X}=(X_1,X_2,X_3,X_4,X_5)$ ($p=5$) and  the 1st tree in the 1st layer uses  $X_3,X_4$ and $X_5$ as split variables, each once at some nodes in the tree, then $\mathbf{c}^{(1)}_{X,1}=(0,0,1,1,1)$.  If $\mathbf{X}^{(l)}$ employs the global X option, then the original $\mathbf{X}$ participates in constructing  trees in each layer. For example, $X_3$ is used twice as the split variable in two nodes, and $X_1$ and $X_5$ once each as the split variable in the 10-th tree in the 2nd layer, then $\mathbf{c}^{(2)}_{X,10}=(1,0,2,0,1)$.  If the local X option is used, then the original $\mathbf{X}$ no longer directly contributes to the construction of trees for layer $l>1$ (that is, $\mathbf{c}^{(l)}_{X,i}\equiv\mathbf{0}$ for $l>2$ in the local X option case), but its contribution will be acknowledged through the selected trees in the previous layer.

We also define $\mathbf{c}_{T,i}^{(k,l)}$ for the layer pair $(l,k)$: $l\ge2$ and $1\le k \le l-1$, which is a column vector of length $m^{(k)}$, containing the frequency of the learned representation attributes from a previous layer $k$ that are selected as split variables for the $i$-th randomized tree in the $l$-th layer. If the local X option is employed, then $\mathbf{c}_{T,i}^{(k,l)}\equiv0$ for $k<l-1$. Imagine that TreeSelection chooses $m^{(1)}=50$ representation features out of $M=500$ randomized trees from layer 1. The 5-th randomized tree in the 2nd layer of SCORE uses 15 representation features, once each, out of the $50$ features. Then  $\mathbf{c}_{T,5}^{(1,2)}$ is a vector of length 50, with 1 in the positions corresponding to the used 15 representation features and 0 elsewhere.

We are now ready to define the VIM for the original  input features $\mathbf{X}$.  If the local X option  is employed, then the VIM for $\mathbf{X}$ is defined as follows:
\begin{align}\label{eqn:vimlocal} 
\!\mbox{VIM}\!=\!\textstyle\sum_{l=1}^L\! \mbox{VIM}^{(l)}\!=\!\sum_{l=1}^L\!\bigg\{\!
\sum_{i_1=1}^{M^{(1)}}\!\sum_{i_2=1}^{M^{(2)}}\!\ldots\!\sum_{i_l=1}^{M^{(l)}}\!|\beta_{i_l}^{(l)}|c^{(l-1,l)}_{T,i_l,i_{l-1}}\!c^{(l-2,l-1)}_{T,i_{l-1},i_{l-2}}
\ldots c^{(1,2)}_{T,i_2,i_1}\!
\mathbf{c}^{(1)}_{X,i_1}\!\!\bigg\}.
\end{align}
$\mbox{VIM}^{(l)}$ quantifies the importance of the original $\mathbf{X}$ in learning the $l$-th layer of the SCORE structure, $c^{(l-1,l)}_{T,i,j}$ denote the $j$-th element in $\mathbf{c}^{(l-1,l)}_{T,i}$, and $M^{(l)}$ refers to the number of representation features selected via the regularized regression in layer $l$.  If the global X option is employed, the calculation of VIM is more complicated as the original and learned representation features not only make the input attributes for the next immediate layer, but possibly also for each layer after that. Hence, the defined VIM has to account for this multiplicity effect. Specifically, we calculate
\vspace{-6pt}
\begin{itemize}[leftmargin=12pt]\setlength\itemsep{-1pt} 
\item the direct contribution of $\X$ to each layer $l$ for $1\le l\le L$
\begin{align}\label{eqn:direct} 
\!\!\!\!\mbox{VIM:direct}=\textstyle\sum_{l=1}^{L}\mbox{VIM}^{(l)}\!=\!
\sum_{l=1}^{L}\sum_{i=1}^{M^{(l)}}|\beta_{i}^{(l)}|\mathbf{c}^{(l)}_{X,i};
\end{align}
\item the indirect contribution of $\X$ through the selected representation features in layer $l$ for $2\le l\le L$
\begin{align}\label{eqn:indirect} 
&\textstyle\mbox{VIM:indirect}=\sum_{l=2}^{L}\mbox{VIM}^{(l)}\notag\\
=&\sum_{l=2}^{L}\sum_{i_1=1}^{M^{(1)}}\!\sum_{i_2=1}^{M^{(2)}}\ldots
\sum_{i_{l-1}=1}^{M^{(l-1)}}\!\sum_{i_l=1}^{M^{(l)}}
\left\{|\beta_{i_l}^{(l)}|
c^{(l,l-1)}_{T,i_l,i_{l-1}}
c^{(l-1,l-2)}_{T,i_{l-1},i_{l-2}}\ldots
c^{(2,1)}_{T,i_2,i_1}
\mathbf{c}^{(1)}_{X,i_1}\right\}.
\end{align}
\end{itemize}
The final VIM for the original $\mathbf{X}$ is defined as the sum of Eqns (\ref{eqn:direct}) and  (\ref{eqn:indirect}),
\begin{equation}\label{eqn:vimglobal}
 \mbox{VIM = VIM:direct + VIM:indirect}
\end{equation}

\section{Experiments}\label{sec:experiments}
We compare SCORE with three ensemble methods -- ExtraTrees, RF, GBM  -- in prediction accuracy, variable importance, and  computational efficiency. In addition, we include NN as another benchmark method due to our use of representation learning. The reasons for choosing the ExtraTrees, RF and GBM for comparison are as follows. First, SCORE builds parallel trees similar to RF and ExtraTrees . Second, SCORE incorporates boosting using functional gradients similar to GBM. Third, ExtraTrees, as one of the base steps for SCORE, epitomizes the use of extreme randomization to maximize diversity among base learners. All compared methods enjoy vast popularity among practitioners for prediction tasks. In addition, all included methods have accompanying VIMs.

\subsection{Experiment Setup}
We run five experiments. The first three  experiments use the Friedman 1, 2, and 3 data sets,  the fourth runs on the real-life Boston Housing data, and the fifth on the real-life World Happiness Report data \cite{Helliwell2019World2019}. 

The outcome $Y$ in Friedman 1 data is simulated as
$Y\!=\!10 \sin( \pi X_1 X_2)\!+\!20( X_3 -0.5)^2+10 X_4 +5 X_5 +\varepsilon$,
where  $X_j\sim U(0,1)$ for $j=1,\ldots,5$ and $\varepsilon \sim N(0,1)$. To examine the efficiency of each method in measuring variable important, we add irrelevant features to the training and testing data. Specifically, we examine two scenarios: 1) 5 additional irrelevant input attributes sampled from $U(0,1)$, totaling to $p=10$; 2) 45 irrelevant input attributes, of which 5 are sampled from $U(0,1)$, 20 from $N(0,1)$, and 20 from Lognormal$(0,1)$, totaling to $p=50$. We examine different training sizes at $10p,50p,100p,200p, 400p$, respectively; and the size of the testing data is fixed at 5,000.  The outcomes in the Friedman 2 and Friedman 3 experiments are simulated from
$Y\!=\!\displaystyle X_1^2+(X_2 X_3 - (X_2 X_4)^{-2})^{0.5} + \varepsilon$ with $\varepsilon \sim N(0,125)$, 
and $Y=\displaystyle\mbox{atan} ((X_2 X_3 - (X_2 X_4)^{-1}))/X_1) + \varepsilon$ with  $\varepsilon \sim N(0,0.1)$, respectively, where $X_1\sim U[0,100], \ X_2\sim U[40\pi,560\pi], \ X_3\sim U[0,1]$, $X_4\sim U[1,11]$ in both cases. Five irrelevant $X$'s are simulated from $N(0,1)$ and added to the training and testing data ($p=9$) in Friedman 2 and Friedman 3 data. Both the training and the test data are of size $n=5,000$.  The Boston housing data contain 506 census tracts of Boston from the 1970 census. The outcome variable is housing price. There are 17 input attributes including housing locations, location characteristics, etc (see Appendix for the list of variables). The world happiness report data (\url{https://worldhappiness.report/ed/2018/}) contain the survey data on the state of happiness. The outcome is the happiness index, with 10 being the best and 0 being the worst possible life. After excluding data with missingness, we include 382 cases. There are 17 input attributes which measure the economic, quality-of-life, and political aspects of different countries (see Appendix for the list of variables). 

R packages \texttt{randomForest, ranger, gbm} were employed to run RF, ExtraTrees, and GBM, respectively; and the R package \texttt{caret} was used to tune the hyper-parameters  with the  10-fold cross validation (for GBM, the maximum number of trees was fixed at 500 in each experiment).  

For NN, the number of hidden layers was set at 2 to match the depth of SCORE for all 5 experiment (for the Friedman 1 experiment, NNs with one hidden layer were also examined but the performance was not good).  R package \texttt{neuralnet} was used to compare the number of hidden nodes per hidden layer,  early stopping rules, and learning rates. The hyper-parameter settings that lead to the highest prediction accuracy on the test data were used: 2 hidden layers with 10 hidden nodes per layer in the Friedman 1 and 2 experiments and the Boston Housing data, 2 hidden layers with 5 hidden nodes per layer in the Friedman 3 experiment and the World Happiness Report data.

For SCORE, we examine 9 hyper-parameter settings at each $n$ in the Friedman 1 experiment (Table \ref{tab:9}) by varying the minimal node size to training size  ratio ($1/5,1/4,1/3$), the number of layers and the number of trees per layer ($L=1$ with 500 trees, $L=2$ with 250 trees per layer), and the global $X$ vs. local $X$ options for $L=2$, in the ExtraTrees step.  For the TreeSelection step in SCORE, we  employed the R package \texttt{glmnet} to run the lasso regression with the built-in  procedure to tune  the hyper-parameter.  
\begin{table}[!htb]
\centering
\caption{Nine SCORE hyper-parameter scenarios in the Friedman 1 experiment}\label{tab:9}
\begin{tabular}{c|c|c|c}
\hline
minimal node size to    & $L=1$  & \multicolumn{2}{c}{$L=2$}\\
\cline{3-4}
training size  ratio  &  & local X option & global X option\\
\hline
 1/5 &  1 & 4 & 7 \\
 1/4 &  2 & 5 & 8\\
 1/3 &  3 & 6 &9\\
 \hline
\end{tabular}
\end{table}
In the other 4 experiments,  rather than tuning the hyper-parameters in each experiment, we examined findings from the Friedman 1 experiment to evaluate how the hyper-parameters impact the performance in  SCORE and  pre-specified the tuning parameters in other experiments based on it. Specifically,  we used a two-layer SCORE structure with the global $X$ option and 5 trees as the minimal node size for Friedman 2 and 3 and the Boston housing data, 25  trees as the minimal node size for the world happiness  data. 

For the VIM measures in ExtraTrees, RF, and GBM, the R function \texttt{varImp} from package \texttt{caret} was applied. 
Though NNs, especially those with deep structures, are known for lacking explainability, there exists some VIMs for  fully-connected feedforward NNs \cite{olden2004accurate}, which are the types of NNs we applied to the five experiments. We applied the Olden method via R package \texttt{NeuralNetTools} to calculate the VIM for the NNs, which measures the relative importance of the input features as the sum of the product of the raw input-to-hidden, hidden-to-output weights.  

Across all 5 methods, the prediction, VIM and computational time results were averaged across 100 repeats. For the Friedman 1, 2, and 3 experiments, we directly simulated 100 sets of training data from the models. For the Boston housing data, we randomly selected 400 cases as the training samples without replacement, and used the rest (106)  as testing samples. The process was repeated 100 times. For the World happiness data,  we randomly selected 250 cases  as the training sample and used the rest (132) as the test samples; and the process was repeated 100 times to generate 100 repetitions.

\subsection{Results on Prediction Accuracy}
\begin{figure}[!htb]
\centerline{\includegraphics[width=1.05\textwidth]{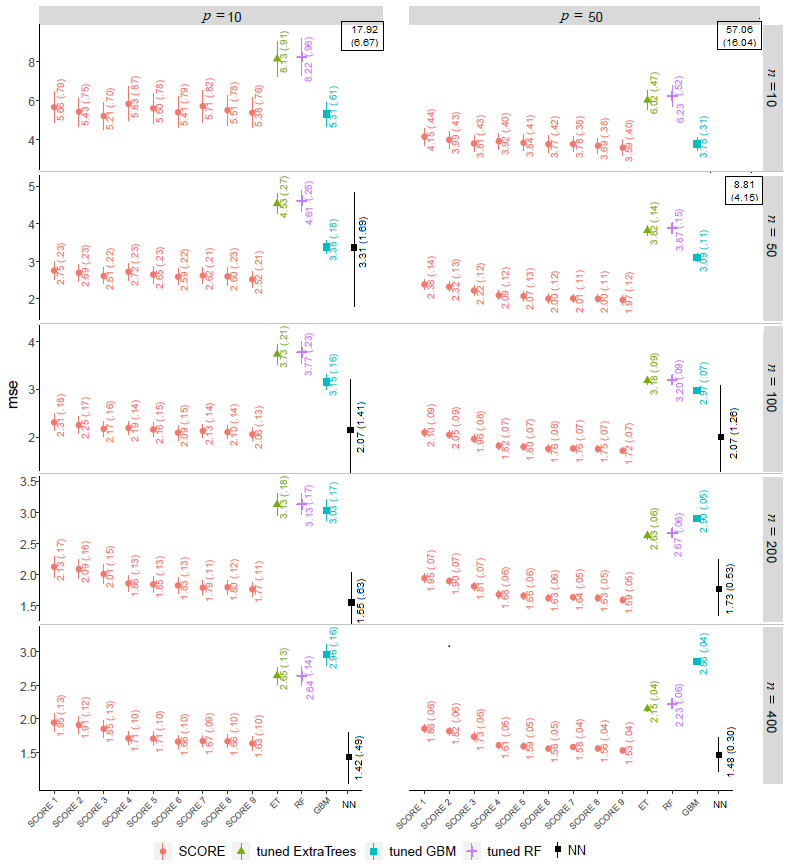}}
\caption{Average ($\pm$ SD) prediction MSE over 100 repeats in Friedman 1 data}
\label{fig:1}
\end{figure}

The prediction results for the Friedman 1 data are displayed in Figure \ref{fig:1}. 
First, the prediction improves in all methods as $n$ increases.   Compared to the 3 ensemble methods, SCORE shows smaller prediction MSE and better performance over ExtraTrees and RF,  and was generally comparable to  GBM across all the experimental settings.  Compared to NN, SCORE is universally better when $p=50$, and for $n=100; 500; 1,000$ when $p=10$; NN is slightly better than SCORE at $n=4,000$ for $p=10$ and $p=50$, at $2,000$ when $p=10$. In sum, good prediction of NN requires a large $n$. In addition, the prediction based on NN is less stable compared to other methods (much larger SD of MSE for all scenarios).  There are some  differences among the 9 SCORE cases with varying hyper-parameter settings, but they are generally comparable. Specifically, having more layers  (SCORE cases 4 to 9) improves the prediction accuracy compared to the single-layered SCORE cases (1 to 3). The minimal node size to the training size ratio seems to affect the predictive accuracy: increasing ratio seems to lead to higher predictive accuracy. This is not unexpected as larger trees tend to over-fit whereas the higher ratio helps control over-fitting. The global $X$ option (cases 7 to 9) offers slight advantages over the local $X$ option (cases 4 to 6). Overall, the small differences among different hyper-parameter settings imply that SCORE is relatively robust to the hyper-parameter specification and can deliver comparable prediction without much tuning.  

Table \ref{tab:23Boston} presents the prediction accuracy for the Friedman 2,  Friedman 3, the Boston housing, and the world happiness experiments. SCORE delivers the smallest MSE in general compared to the three 3 approaches.
\begin{table}[!htb]
\centering
\caption{Mean Prediction MSE (SD) in Experiments 2 to 5}\label{tab:23Boston}
\begin{tabular}{@{}c@{\hskip 3pt}| @{\hskip 3pt}c@{\hskip 3pt}| @{\hskip 3pt}c@{\hskip 3pt}| @{\hskip 3pt}c@{\hskip 3pt}|
@{\hskip 3pt}c@{}}
\hline
& Friedman 2 &  Friedman 3 &  Boston  & happiness\\
&& & housing &  report\\
\hline
SCORE & $\mathbf{16.6\times10^3}$ (418.3) & $\mathbf{0.0119}\;(3.443\times10^{-4}$) & \textbf{9.0} (2.1) & \textbf{0.07} (0.015) \\
ExtraTrees & $17.1\times10^3$ (367.8) & $0.0120\;(3.235\times10^{-4}$) & 13.0 (4.0) & 0.08 (0.014)\\
RF & $17.1\times10^3$ (360.7) & $0.0120\; (3.197\times10^{-4}$) & 10.3 (3.5) & \textbf{0.07} (0.016) \\
GBM & $35.4\times10^3$ (728.2) & $0.0236\;(6.031\times10^{-4}$) & 11.8 (3.2) & 0.11 (0.018)\\
NN & $21.5\times10^3$ (4277.0) & $0.0127 \ (2.280 \times 10^{-3}$) & 16.2 (5.) & 0.10 (0.034)\\
\hline
\end{tabular}
\end{table}

\subsection{Results on Variable Importance}
The list of the identified important variables are presented  in Table  \ref{tab:impvar}, based on the average normalized VIM across the 100 repetitions.  We  applied $1/p$ as the cutoff to identify the relevant attributes in all methods. 
\begin{table}[!htb]
\centering
\caption{Identified Relevant Input Attributes}\label{tab:impvar}
\resizebox{1\columnwidth}{!}{%
\begin{tabular}{@{}l|@{\hskip 1pt}l|@{\hskip 1pt}l|@{\hskip 1pt}l|@{\hskip 1pt}l|@{\hskip 1pt}l@{}}
\hline
Data & \multicolumn{3}{c|}{Friedman} & Boston & happiness \\
\cline{2-4}
 $(1/p)^\ddagger$ & 1 ($p\!=\!50; n\!=\!5,000; 2\%$) & 2 (11.1\%) & 3 (11.1\%) & housing (5.88\%) & report (5.88\%)\\
 \hline
SCORE & $X_2, X_1, X_3, X_5, X_4$ & $X_3,X_2$ & $X_3,X_1,X_2$ & rm, lstat, age, crim, lon, & sdml, sdl, lGDPpc, \\
&&&& b, dis, tract, lat, nox & life,  posaf,  del, sup \\
ET & $X_4, X_2, X_1, X_5, X_3$ & $X_3,X_2$  & $X_3, X_2, X_1$ & rm, lstat & sdml, del, life,\\
&&&& 
& lGDPpc,  posaf\\
RF & $X_4, X_2, X_1, X_5, X_3$ & $X_3,X_2$  & $X_3,X_1,X_2$ & lstat, rm  & sdml, life, sdl \\
&&&& \\ 
GBM & $X_4, X_2, X_1, X_3, X_5$ & $X_3,X_2$  & $X_3, X_2, X_1$& lstat, rm & sdml, life, posaf, \\ 
&&&& 
& lGDPpc 
\\
NN & $X_2, X_1, X_3, X_4, X_5,X_8$ & $X_1$ & $X_3,X_1,X_2$ & lstat, dis, rm, nox, & sdml, sdl  \\
&  &  &  &  crim, tax, tract & 
\\
\hline
\end{tabular}}
\begin{tabular}{l}
\footnotesize The attributes are listed in the order of descending VIM.\textcolor{white}{.\hspace{2.8in}.}\\
\footnotesize $^\ddagger$ $>1/p$ is used as the cutoff for normalized VIM to identify relevant attributes. \\
\hline
\end{tabular}
\end{table}

In the Friedman 1 experiment, all approaches succeeded in identifying $X_1$ to  $X_5$ as the five most important features  and the other 45 variables (which we added artificially) as irrelevant.  For SCORE and ExtraTrees, $X_1$ to $X_5$ had considerably larger weights than the 45 irrelevant features, but ExtraTrees put much less weight on  $X_3$ (3.6\%)  compared to SCORE (11.5\%).  

In the Friedman 2 experiment,  all the methods except for NN selected $X_2$ and $X_3$ as the top two attributes while NN picked $X_2$. The sub-optimal performance is likely due to the difficulty associated with the underlying model. The true underlying model that is simulated to simulate the data shows that $X_2$ and $X_4$ appear together as in the product term $(X_2X_4)^{-2}$, which can be very small given the large magnitude of $X_2$, resulting in a smaller contribution toward $Y$ relative to others terms. In other words, the large VIM for $X_2$ likely comes from the $X_2X_3$ term. The failure to select  $X_1$ is likely due to the the large spread of $X^2_1$ ($0$ to $10^4$) compared to the standard deviation of the error term, making the signals from $X_1$ indistinguishable from the background noise.  

In the Friedman 3 experiment,  all methods successfully identified $X_1,X_2,X_3$ as relevant variables but missed $X_4$. Though $X_4$ is ranked in the 4-th place in SCORE and ExtraTrees, the corresponding VIM is only slightly higher than for the rest. The failure for identifying $X_4$ in this case is likely due to the same reason listed for Friedman 2 data. 

For the Boston housing data, all methods agree that \textit{rm} (the number of rooms) and \textit{lstat} (proportion of people with lower socioeconomic status) are two important variables for predicting the housing price. For the next 3 important variables, SCORE identified \textit{age} (proportion of owner-occupied units built prior to 1940), \textit{crim} (per capita crime rate by town), and  \textit{lon} (longitude of census tract); ExtraTrees identified -- not listed in Table \ref{tab:impvar} -- \textit{ptratio} (pupil-teacher ratio by town), \textit{tax}(full-value property-tax rate per USD 10,000), and \textit{indus}(proportion of non-retail business acres per town); RF identified \textit{nox} (nitric oxides concentration), \textit{crim} (per capita crime rate by town), and \textit{dis} (weighted distances to 5 Boston employment centres); GBM identified \textit{tract} (census tract), \textit{town} (name of the town), \textit{dis}; and NN identified dist, nox, and crim. The 10 important predictors identified by SCORE included almost all the top-five predictors identified by RF and NN. Note that \textit{tract} (census tract), \textit{town} (name of the town), and (\textit{lon}, \textit{lat}) (longitude \& latitude of a census tract) containing about the same information, so identifying one or the other does not matter much from an interpretation perspective. 

For the world happiness report data, all methods agree that \textit{sdml} (standard deviation/mean of the ladder index) is the most important predictor for happiness. For the next 4 important variables, SCORE identified \textit{sdl} (standard deviation of the ladder index), \textit{lGDPpc} (log GDP per capita), \textit{life} (healthy life expectancy at birth), and \textit{posaf} (positive affect); ExtraTrees identified \textit{del} (delivery quality), \textit{life} (healthy life expectancy at birth), \textit{lGDPpc} (log GDP per capita), \textit{posaf} (positive affect); RF identified \textit{sdml} (standard deviation/mean of the ladder index), \textit{life} (healthy life expectancy at birth), \textit{sdl} (standard deviation of the ladder index), \textit{posaf} (positive affect), and \textit{lGDPpc} (log GDP per capita); GBM identified, \textit{sdml} (standard deviation/mean of the ladder index), \textit{life} (healthy life expectancy at birth), \textit{posaf} (positive affect), \textit{lGDPpc} (log GDP per capita), and \textit{sup} (social support); NN identifies \textit{sdl} (standard deviation of the ladder index), \textit{gini2} (GINI index Word Bank estimate), \textit{gini1}(GINI index Word Bank estimate, average 2000-2015), \textit{del} (delivery quality). SCORE and RF agree on all top five important variables. 

In summary, the VIM procedure associated with SCORE shows good performance in identifying important predictors and in general agrees with RF and GBM on the important variable list in these 5 experiments.

\subsection{Computational Time}
We present the computational time from the Friedman 1 experiment (Table \ref{tab:time}) (averaged over 100 repeats). The observations are similar in other experiments. The time spent on tuning hyper-parameters via the 10-fold CV is included in all methods except for SCORE and NNs, where the hyper-parameters were pre-specified by comparing a limited set of scenarios. All methods were run in R version 3.5.3 on 12 core Intel(R) Haswell processors with 256 GB RAM. In summary, SCORE was the fastest in general. The speed of NN is greatly affected by the stopping criterion used (early stopping can save a significant amount of time). The computational advantage for SCORE is particularly evident for the larger $n$ case. Moreover, the computational time for two-layered SCORE (cases 4 to 9) with 250 trees per layer is less than the singled-layered SCORE (cases 1 to 3) with 500 trees. The computational time also decreased as the minimal node size to training size  ratio  increases as the tree stops growing earlier with a larger ratio. The global $X$ options, as using more input attributes, was more time-consuming than the local options.
\begin{table}[!htb]
\centering
\caption{Average Computational Time (seconds) in Friedman 1 Data}\label{tab:time}
\resizebox{1\textwidth}{!}{
\begin{tabular}{r| rrrrrr r}
 \hline
\textcolor{white}{.\hspace{0.6in}.}& $n$ &  100 &  500 &  1,000 &  2,000 & 4,000 & \textcolor{white}{.\hspace{0.6in}.} \\
 \cline{2-8}
& SCORE case 1&0.75 &2.09 &2.28 &4.25 & 9.61 \\
& SCORE case 2&0.74 &2.20 &2.32 &4.30 & 10.39 \\
& SCORE case 3&0.77 &2.46&2.44 &4.28 & 10.11 \\
& SCORE case 4&0.55 &1.92 &6.55 &19.21 & 43.78 \\
$p=10$ & SCORE case 5&0.51 &1.92 &7.00 &21.77& 49.54 \\
& SCORE case 6&0.50 &1.97 &7.96 &24.03 & 59.87 \\
& SCORE case 7&0.77 &2.12 &2.34 &4.34 & 9.48 \\
& SCORE case 8&0.75 &2.21 &2.36 &4.34 & 10.34 \\
& SCORE case 9&0.77 &2.44 &2.45 &4.34 & 10.34 \\
& ExtraTrees&12.89&14.52&26.29&79.40& 168.72 \\
& RF&10.07&79.26&204.00&754.71& 1830.51 \\
& GBM&51.71&64.88&68.54&126.27& 145.81 \\
& NN & 0.38 & 8.43 &   21.23 &  75.20  &  91.44\\
 \hline
& $n$ &  500 & 2,500 & 5,000 & 10,000 & 20,000 \\
 \cline{2-8}
& SCORE case 1&2.96&5.63&14.94&42.7& 117.10 \\
& SCORE case 2&3.23&5.50 &14.94&48.58& 149.09 \\
& SCORE case 3&3.71&5.55&15.26&63.14& 180.27 \\
& SCORE case 4&2.78&20.13&60.37&143.4& 387.52 \\
$p=50$& SCORE case 5&2.85&21.85&63.99&165.95& 521.57 \\
& SCORE case 6&2.98&23.46&71.08&215.01& 721.38 \\
& SCORE case 7&2.92&5.66&15.11&42.82& 116.46 \\
& SCORE case 8&3.10 &5.62&15.56&48.74& 149.47 \\
& SCORE case 9&3.42&5.62&15.63&58.77& 183.54 \\
& ExtraTrees&79.18&228.78&470.48&1862.43& 4462.34 \\
& RF&461.02&3967.89&13,084.36&32,237.32& 10,6692.39 \\
& GBM&1,023.23&1,458.58&291.24&528.53&807.90\\
& NN$^*$ & 1.96 &  73.17& 75.67 & 50.45 & 127.47\\
\end{tabular}}
\begin{tabular}{p{1\textwidth}}
\hline
\footnotesize $L=1$ in SCORE 1 to  3 with 500 trees; $L=2$ in SCORE 4 to 9 with 250 trees per layer; the minimal node size to training size  ratio is 1/5 for SCORE 1, 4, 7, 1/4 for SCORE 2, 5, 8; and 1/3 for SCORE 3, 6, 8;  SCORE 4 to 6 use the local X option; and SCORE 7 to 9 employ the global X option. Refer to Table \ref{tab:9}.\\
\footnotesize $^*$ The computational time does not increase with $n$ for NN because the stopping criteria, which is the threshold for the partial derivatives of the error function, are different across different $n$. Larger errors were allowed for larger $n$ due to non-convergence issues. Specifically, the error was set at 0.01 (default in the \texttt{neuralnet} function), 0.025, 0.04, 0.1, 0.3, and 0.4 for the 5 $n$, respectively.\\
\hline
\end{tabular}
\end{table}

\section{Discussion}
We have proposed a novel tree ensemble method, Selective  Cascade of Residual ExtraTrees (SCORE), for regression prediction. SCORE incorporates a TreeSelection step that weighs each tree based on its relevance to the outcome prediction. This step filters out the irrelevant trees, which  helps reduce not only the bias but also the variance for prediction without hurting the diversity among the ensemble members.  Moreover, the weights obtained through the  TreeSelection step can be used to evaluate the importance of predictors involved in building each tree, on which our new VIM is based. The boosting step in SCORE helps to further reduce the bias from the potential over-randomization in ExtraTrees. The experiments show that SCORE offers comparable or superior performance in prediction and predictor selection compared to ExtraTrees, RF, and GBM at significantly less computational time, owing to its robustness to hyper-parameter specifications. Similar observations are obtained comparing SCORE to NNs, suggesting NNs are not universally better than other leaning approaches, depending on the specific data as well as training data sizes. Moreover, compared to other tree-based ensemble methods \cite{Zhou2017DeepNetworks,Feng2018Multi-layeredTrees, Kong}, SCORE's ability to measure variable importance as relevant to the prediction process offers  explainability on its prediction, and  provides transparency of and insight into the modeling processes.

Future research may extend the framework of SCORE to classification tasks. The investigation on the robustness of SCORE to hyper-parameter specifications in terms of prediction performance also awaits a comprehensive investigation.

\section*{Acknowledgement}
We thank Dr. Gitta Lubke and two anonymous referees for their useful and constructive comments on the project and the manuscript. 

\bibliographystyle{IEEEtran}
\bibliography{ref.bib}

\section*{Appendix}
\centering
Input Attributes in Boston Housing Data
\resizebox{0.7\textwidth}{!}{
\begin{tabular}{ll}
\hline
label & attribute\\
\hline
\textit{crim} &	per capita crime rate by town\\
\textit{zn} &	proportion of residential land  zoned for lots over  \\ 
& 25,000 sq.ft\\
\textit{indus}&	proportion of non-retail business acres per town\\
\textit{chas}&	tract bounds Charles River? \\
\textit{nox}&	nitric oxides concentration (parts per 10 million)\\
\textit{rm}&	average number of rooms per dwelling\\
\textit{age}&	proportion of owner-occupied units built prior to 1940\\
\textit{dis}&	weighted distances to five Boston employment centres\\
\textit{rad}&	index of accessibility to radial highway\\
\textit{tax}&	full-value property-tax rate per USD $10,000$\\
\textit{ptratio}&	pupil-teacher ratio by town\\
\textit{b}&   $1000(B-0.63)^2$, where $B$ is the proportion of \\
& blacks by town\\
\textit{lstat}&	percentage of lower status of the population\\
\textit{town}&	name of town\\
\textit{tract}&	census tract \\
\textit{lon}&	longitude of census tract\\
\textit{lat}&	latitude of census tract\\
\hline
\end{tabular}}

\centering
Input Attributes in World Happiness Report Data
\resizebox{0.7\textwidth}{!}{
\begin{tabular}{ll}
\hline
label & attribute\\
\hline
\textit{yr} &	year\\
\textit{lGPDpc} &	logged GDP per Capita\\
\textit{sup}& social support\\
\textit{life}&	healthy life expectancy at birth \\
\textit{free}&	freedom to make life choices\\
\textit{genr}&	generosity\\
\textit{corr}&	perceptions of corruption\\
\textit{posaf}&	positive affect\\
\textit{negaf}&	negative affect\\
\textit{govcon}& confidence in national government\\
\textit{dem}&	democratic quality\\
\textit{del}&  delivery quality\\
\textit{sdl}&	standard deviation of ladder by country-year\\
\textit{sdml}&	standard deviation/mean of ladder by country-year\\
\textit{gini}&	GINI index (World Bank estimate) \\
\textit{gini2}&	GINI index (World Bank estimate), average 2000-2015\\
\textit{gini3}&	GINI of household income reported in Gallup world \\
& poll, by country-year\\
\hline
\end{tabular}}
\end{document}